\documentclass[10pt,twocolumn,letterpaper]{article}

\usepackage[pagenumbers]{cvpr} %
\usepackage[accsupp]{axessibility} %

\usepackage[dvipsnames]{xcolor}

\newcommand{\unipara}[1]{\noindent \textbf{#1}}

\definecolor{cvprblue}{rgb}{0.21,0.49,0.74}
\usepackage[pagebackref,breaklinks,colorlinks,citecolor=cvprblue]{hyperref}

\usepackage{colortbl}
\usepackage{booktabs,multirow}
\usepackage{makecell}
\usepackage{tabulary}
\usepackage{fontawesome5}
\newcommand \blfootnote[1]{
    \begingroup
        \renewcommand
        \thefootnote{}\footnote{#1}
        \addtocounter{footnote}{-1}
        \vspace{-1ex}
    \endgroup
}
\newlength\savewidth\newcommand\shline{\noalign{\global\savewidth\arrayrulewidth
  \global\arrayrulewidth 1pt}\hline\noalign{\global\arrayrulewidth\savewidth}}

\newcommand{\tablestyle}[2]{\setlength{\tabcolsep}{#1}\renewcommand{\arraystretch}{#2}\centering\footnotesize}
\newcommand{\fullmethod}{Action-Disentangled Identifier\xspace}
\newcommand{\method}{ADI\xspace}
\newcommand{\benchmark}{ActionBench\xspace}

\def\paperID{7052} %
\def\confName{CVPR}
\def\confYear{2024}

\title{Learning Disentangled Identifiers for Action-Customized \\Text-to-Image Generation}

\author{Siteng Huang$^{1,4}$\textsuperscript{*}, Biao Gong$^{2}$, Yutong Feng$^{2}$, Xi Chen$^{2}$, Yuqian Fu$^{3}$, Yu Liu$^{2}$, Donglin Wang$^{4}$\footnotemark[2]\>\,\\
{$^1$Zhejiang University\ \ $^2$Alibaba Group\ \ $^3$ETH Zürich}\\
{$^4$Machine Intelligence Lab (MiLAB), AI Division, School of Engineering, Westlake University}\\[-2pt]
{\tt\small \{siteng.huang, a.biao.gong, yuqianfu0207\}@gmail.com}\\[-3pt]
{\tt\small \{fengyutong.fyt, xizhi.cx, ly103369\}@alibaba-inc.com}\ \ {\tt\small wangdonglin@westlake.edu.cn}
}

\begin{document}

\twocolumn[{
\maketitle
\vspace{-4mm}
\begin{center}
    \captionsetup{type=figure}
    \includegraphics[width=\linewidth]{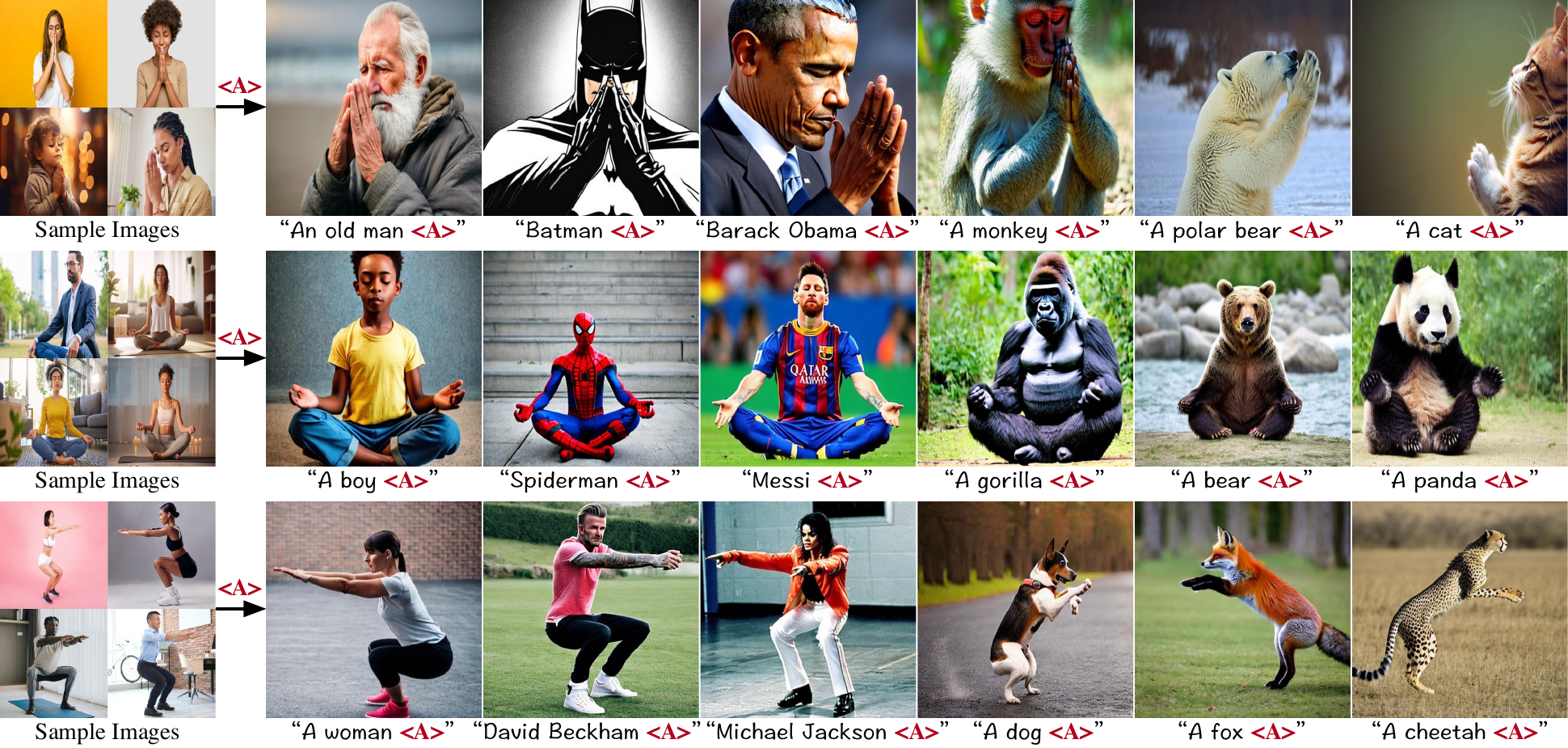}
    \vspace{-6mm}
    \captionof{figure}{\textbf{Action customization results of our \method method.} 
    By inverting representative action-related features, the learned identifiers ``$<\!\!\text{A}\!\!>$'' can be paired with a variety of characters and animals to contribute to the generation of accurate, diverse and high-quality images.
    }
    \label{fig:teaser}
\end{center}
}]

\blfootnote{\textsuperscript{*}Work done during internship at Alibaba Group.}
\blfootnote{$^\dagger$Corresponding author.}

\begin{abstract}

This study focuses on a novel task in text-to-image (T2I) generation, namely \textbf{action customization}.
The objective of this task is to learn the co-existing action from limited data and generalize it to unseen humans or even animals.
Experimental results show that existing subject-driven customization methods
fail to learn the representative characteristics of actions and 
struggle in decoupling 
actions from context features, including appearance.
To overcome the preference for low-level features and the entanglement of high-level features,
we propose an inversion-based method \textbf{\fullmethod (\method)} to learn action-specific identifiers from the exemplar images.
\method first expands the semantic conditioning space by introducing layer-wise identifier tokens, 
thereby increasing the representational richness
while distributing the inversion across different features.
Then, to block the inversion of action-agnostic features, \method extracts the gradient invariance from the constructed sample triples and masks the updates of irrelevant channels.
To comprehensively evaluate the task, we present an \textbf{\benchmark} that includes a variety of actions, each accompanied by meticulously selected samples.
Both quantitative and qualitative results show that our \method outperforms existing baselines in action-customized T2I generation. Our project page is at \url{https://adi-t2i.github.io/ADI}.
\vspace{-5mm}
\end{abstract}

\begin{figure*}[!t]    %
  \centering
   \includegraphics[width=0.85\linewidth]{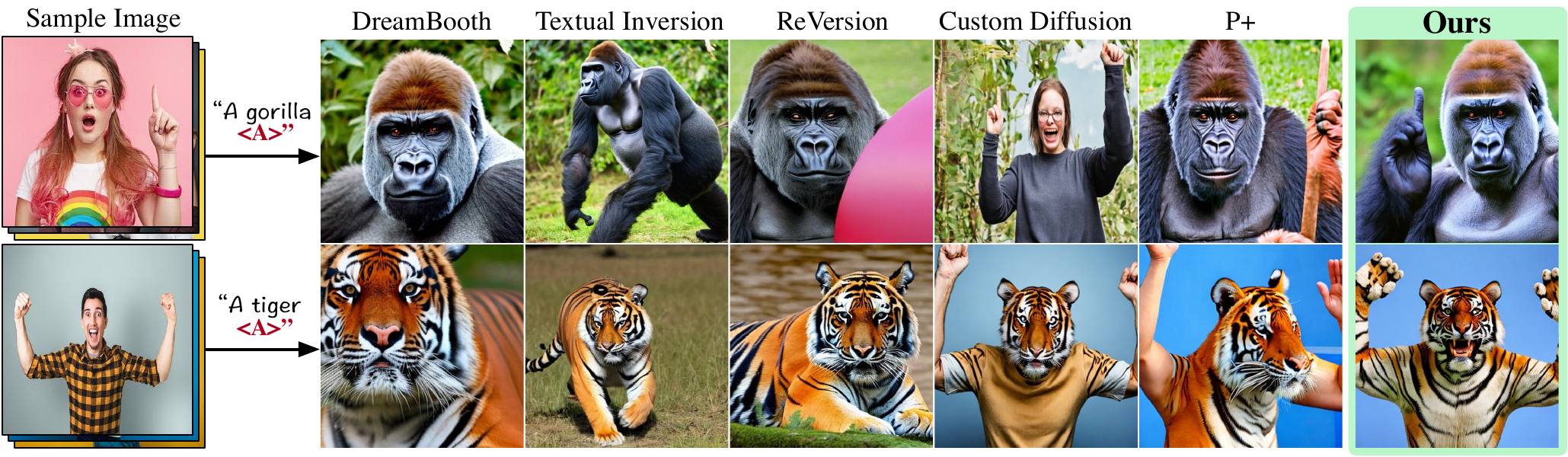}
   \vspace{-2mm}
   \caption{
   \textbf{Action customization results of existing subject-driven customization methods.}
   Due to the preference to search for low-level invariants when asked to learn high-level action features, some methods fail to generate the specified actions, while others confuse the animals with human appearances.
   }
      \vspace{-4mm}
   \label{fig:method_analysis}
\end{figure*}

\section{Introduction}
\label{sec:intro}

Thanks to the remarkable advances in text-to-image generation models~\cite{Reed:GAN-T2I,Ding:CogView,Li:GLIGEN,Ramesh:DALL-E}, in particular the recent diffusion model~\cite{Rombach:Stable-Diffusion,Saharia:Imagen}, high-quality and diverse images can be synthesized under the control of text descriptions.
However, it is difficult to provide precise descriptions of the desired actions, which are highly abstracted and summarized concepts.
Therefore, relying solely on textual descriptions to generate actions tends to reduce fidelity to user requirements.
Additionally, controllable generation methods~\cite{Zhang:ControlNet,Mou:T2I-Adapter} that 
rely on the conditioning of a skeleton or sketch image suffer from limited diversity and freedom, 
and they show difficulty generalizing to unseen subjects without retraining.
In this paper, we study the \textbf{action customization} task, capturing the common action in the given images to generate new images with various new subjects.

To better understand the challenge of action customization,
we start by examining existing subject-driven customization methods.
Observations shown in \cref{fig:method_analysis} can be divided into two categories.
Several methods including DreamBooth~\cite{Ruiz:DreamBooth}, Textual Inversion~\cite{Gal:textual-inversion}, and ReVersion~\cite{Huang:ReVersion}, generate images that are unrelated to specific actions,
suggesting that they fail to capture the representative characteristics of the actions.
Since most of them are designed to invert appearance features with a pixel-level reconstruction loss, low-level details are emphasized during optimization while high-level action features are neglected.
Benefiting from fine-tuning cross-attention or utilizing per-layer tokens, Custom Diffusion~\cite{Kumari:Custom-Diffusion} and P+~\cite{VoynovL:P-Plus} offer a larger semantic conditioning space for learning new concepts.
Consequently, they are capable of encoding action-related knowledge such as ``\textit{raises one finger}'' or ``\textit{raises both arms for cheering}'' from exemplar images.
However, they fail to decouple the focus from action-agnostic features, such as the appearance of the human body.
These pieces of information are also encoded into the learned identifiers and ``contaminate'' the generation of animals during inference.
As a result, the intended gorilla is replaced by a woman, and the tigers generated by the two methods exhibit human arms instead.

To avoid the appearance leakage while accurately modeling the target action, we propose \textbf{\fullmethod (\method)} to learn the optimal action-specific identifiers.
Firstly, we expand the semantic conditioning space by applying layer-wise identifier tokens.
Since existing works have analyzed that different layers have varying degrees of control over low-level and high-level features~\cite{VoynovL:P-Plus}, such an expansion increases the accommodation of various features, making it easier to invert action-related features.
Furthermore, we would like to decouple the action-agnostic features from the learning of action identifiers.
To achieve this, we discover invariant mechanisms in the data that are difficult to vary across examples.
Specifically, given an exemplar image with the specific action, another same-action image can be randomly sampled from the training data, forming a context-different pair.
Meanwhile, leveraging mature subject-driven customization techniques, an image that shares the similar context can be quickly synthesized to form an action-different pair.
To decouple the highly-coupled features, we disentangle action-agnostic features at the gradient level, and construct two context gradient masks by comparing the difference on the gradients over the input pairs.
By overwriting the merged gradient mask to the gradient of the anchor image, the update of action-agnostic channels on the identifiers is discarded.

Moreover, as a pioneering effort in this direction, we also contribute to a new benchmark named \textbf{\benchmark}, which provides a testbed of unique actions with diverse images for the under-explored task.
We conduct extensive experiments on \benchmark, and a quick glance at the performance of \method is illustrated in \cref{fig:teaser}, where users can freely combine the designated action identifiers with various unseen humans and even animals.
In summary, the main contributions of our work are three-fold: %
\begin{itemize}[labelsep=0.4em, leftmargin=1em,itemindent=0em]
	\item We propose a novel action customization task, which requires learning the desired action from limited data for future generation.
    While existing customization focuses on reprinting appearances, we highlight this under-studied but important problem.
	\item We contribute the \benchmark, where a variety of unique actions with manually filtered images provide the evaluation conditions for the task. 
    \item We devise the \fullmethod (\method) method, which successfully inverts action-related features into the learned identifiers that can be freely combined with various characters and animals to generate high-quality images.
\end{itemize}
\section{Related Work}
\label{sec:related}

\unipara{Text-to-Image (T2I) Generation.}
Generating high-quality and diverse images from textual conditions has received considerable attention from both the research community and the general public.
The previous dominant generative adversarial networks (GANs)~\cite{Reed:GAN-T2I,Xu:AttnGAN,Tao:DF-GAN,Zhu:DM-GAN}, consisting of a generator and a discriminator, suffer from unstable optimization and less diverse generations due to the adversarial training~\cite{Dhariwal:diffusion-beats-GAN}.
And variational autoencoders (VAEs)~\cite{Kingma:VAE}, which apply a probabilistic encoder-decoder architecture, are also prone to posterior collapse and over-smoothed generations~\cite{Zhao:InfoVAE}.
Text-conditional auto-regressive models~\cite{Ramesh:DALL-E,Gafni:Make-A-Scene,Ding:CogView,Yu:Parti} have shown more impressive results, but require time-consuming iterative processes to achieve high-quality image sampling.
More recently, diffusion models have emerged as a promising alternative, achieving impressive results with open-vocabulary text descriptions through their natural fitting to inductive biases of image data~\cite{Saharia:Imagen,Ramesh:DALL-E2,Nichol:GLIDE,Rombach:Stable-Diffusion}.
GLIDE~\cite{Nichol:GLIDE} introduces text conditions into the diffusion process through the use of an unclassified guide.
DALL-E 2~\cite{Ramesh:DALL-E2} employs a diffusion prior module and cascading diffusion decoders to generate high-resolution images based on the CLIP~\cite{Radford:CLIP} text encoder.
Imagen~\cite{Saharia:Imagen} focuses on language understanding by using a large T5 language model to better represent semantics.
The latent diffusion model~\cite{Rombach:Stable-Diffusion} improves computational efficiency by performing the diffusion process in low-dimension latent space with an autoencoder.
Finally, Stable Diffusion (SD)~\cite{Rombach:Stable-Diffusion} employs a cross-attention mechanism to inject textual conditions into the diffusion generation process, aligning with the provided textual input.
However, it is difficult to provide precise action descriptions in text, 
since user intent and machine understanding are not aligned.
Furthermore, experimental results in \cref{fig:baselines} show that some actions are difficult to generate correctly without re-training, \textit{e.g.}, ``\textit{performs a handstand}''.

\unipara{Controllable Action Generation.}
The paper focuses on transferring the desired action from examplar images to unseen people, characters, and even animals for photorealistic image generation.
Existing efforts take source images and pose information (\textit{e.g.}, skeletal images or body parsing) as conditions to control the generation.
Previous controllable solutions based on GANs~\cite{Ma:pose-guided-HIG,Men:HIG,Zhang:pose-guided-HIG} and VAEs~\cite{Ren:HIG,Yang:HIG} suffer from training difficulties and poor generation results.
Some subsequent works~\cite{Roy:TIPS,Xu:text-guided-HIG} introduce text conditions to guide the action generation, yet fail with open vocabulary due to the small size of the vocabulary pools.
Thanks to the significant advances of T2I diffusion models, recent methods~\cite{Tumanyan:plug-and-play,Li:GLIGEN,Mou:T2I-Adapter}, in particular the popular ControlNet~\cite{Zhang:ControlNet}, add arbitrary conditions to improve the versatility and controllability.
While gaining a tremendous amount of traction from the community, ControlNet refers to the provided skeleton image to generate the action, which reduces flexibility and diversity.
In addition, the objective of designing a general framework with additional trainable modules makes it not well-targeted to animals.
In this work, we investigate customization solutions for action generation.

\unipara{Subject-Driven Customization.}
Due to the demand for generating images with user-specified subjects, customization methods~\cite{Ruiz:DreamBooth,Gal:textual-inversion,Kumari:Custom-Diffusion,VoynovL:P-Plus} tailored to the appearance have been studied in the context of T2I generation.
Specifically,
DreamBooth~\cite{Ruiz:DreamBooth} binds rare new words with specific subjects through fine-tuning the whole T2I generator.
Textual Inversion~\cite{Gal:textual-inversion} learns an extra identifier to represent the subject and adds the identifier as a new word to the dictionary of the text encoder.
Custom Diffusion~\cite{Kumari:Custom-Diffusion} only fine-tunes the 
key and value matrices
of the cross-attention to represent new concepts.
P+~\cite{VoynovL:P-Plus} extends the textual-conditioning space with per-layer tokens to allow for greater disentangling and control.
Despite the success achieved, the experimental results in \cref{fig:method_analysis} show their failure in action customization.
A recent work ReVersion~\cite{Huang:ReVersion} makes progress in learning specific relations including some interactions from exemplar images.
However, the design of the method, which specializes in learning spatial relations, makes it difficult to invert action information.

\begin{figure*}[htbp]
  \centering
   \includegraphics[width=0.98\linewidth]{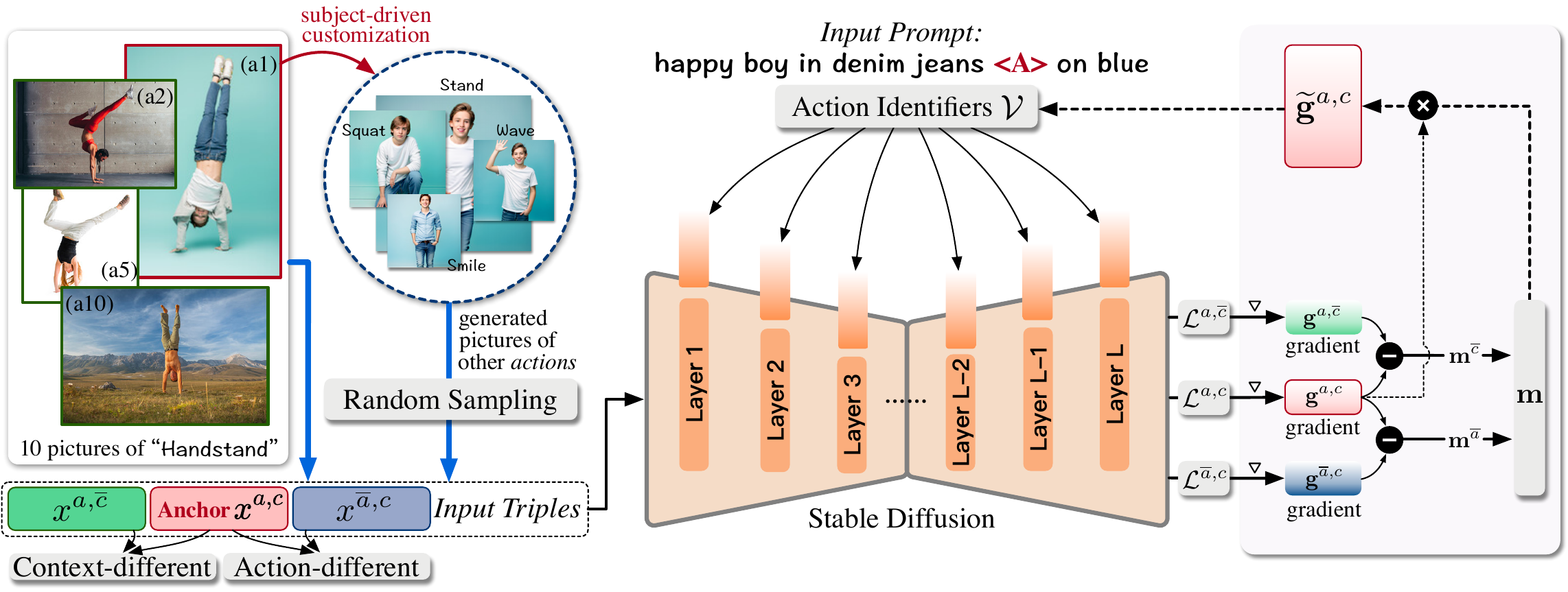}
   \vspace{-2mm}
   \caption{
   \textbf{Overview of our \method method}.
   \method learns more efficient action identifiers by extending the semantic conditioning space and masking gradient updates to action-agnostic channels.
   }
         \vspace{-3mm}
   \label{fig:method}
\end{figure*}

\section{Action Customization Benchmark}

Given a set of exemplar images $\mathcal{X} = \left\{ \mathbf{x}_1, \mathbf{x}_2, \cdots, \mathbf{x}_N \right\}$, we assume that all images contain the same action performed by different people.
The action-agnostic descriptions associated to the exemplar images are also provided, which can be used as prompt templates during training.
The objective of the action customization task is to extract the co-existing action and transfer it to the synthesis of action-specific images with different new subjects.
In order to provide suitable conditions for systematic comparisons on this task, we present a new \textbf{\benchmark}, which consists of diverse actions accompanied by meticulously selected sample images.
The benchmark can be used for both quantitative and qualitative comparisons. 

\unipara{Action Categories.}
To determine the involved actions, 
we first request GPT-4~\cite{OpenAI:GPT-4} to provide 50 candidate action categories, and then attempt to collect images for these candidates.
Only actions that can collect sufficient high-quality images are preserved.
We finally define eight unique actions, ranging from single-handed (\textit{e.g.}, ``\textit{raises one finger}'') to full-body movements (\textit{e.g.}, ``\textit{performs a handstand}'').

\unipara{Exemplar Images and Prompts.}
For each action, we collect ten example images with corresponding textual descriptions, featuring different people.
We manually remove action-related descriptions from the textual content to make them suitable as prompt templates.

\unipara{Evaluation Subjects.}
We provide a list containing 23 subjects, including generic humans (\textit{e.g.}, ``\textit{An old man}''), well-known personalities (\textit{e.g.}, ``\textit{David Beckham}''), and animals (\textit{e.g.}, ``\textit{A panda}'').
The latter two are guaranteed to be completely unseen, which tests the generalization of methods.

\section{Methodology}

We start with the technical background in \cref{sec:preliminaries}.
Then, we provide a comprehensive description of our proposed \method in \cref{sec:method}.

\subsection{Preliminaries} \label{sec:preliminaries}

Our study is based on the Stable Diffusion (SD)~\cite{Rombach:Stable-Diffusion} model, which is considered to be the public state-of-the-art text-to-image generator.
Specifically, to operate the diffusion process~\cite{Ho:diffusion} in a low-dimensional latent space, SD employs a hierarchical VAE that consists of an encoder $\mathcal{E}$ and a decoder $\mathcal{D}$. 
The encoder $\mathcal{E}$ is tasked with encoding the given image $x$ into latent features $\mathbf{z}$, and the decoder $\mathcal{D}$ reconstructs the image $\widehat{x}$ from the latent, \textit{i.e.}, $\widehat{x} = \mathcal{D}(\mathbf{z}) = \mathcal{D}(\mathcal{E}(x))$.
To control the generation with the textual conditions, given the noisy latent $\mathbf{z}_t$, current time step $t$ and text tokens $\mathbf{y}$, a conditional U-Net~\cite{Ronneberger:U-Net} denoiser is trained to predict the noise $\epsilon$ added to the latent $\mathbf{z}$:
{\setlength\abovedisplayskip{2mm}
\setlength\belowdisplayskip{2mm}
\begin{equation}
\mathcal{L}=\mathbb{E}_{\mathbf{z} \sim \mathcal{E}(x), \mathbf{y}, \epsilon \sim \mathcal{N}(0,1), t}\left[\left\|\epsilon-\epsilon_\theta\left(\mathbf{z}_t, t, \mathbf{y}\right)\right\|_2^2\right],
\end{equation}}%
where $\mathbf{y}$ is obtained by feeding the prompt into a CLIP~\cite{Radford:CLIP} text encoder.
During inference, the pre-trained SD first samples a latent $\mathbf{z}_T$ from the standard normal distribution $\mathcal{N}(0,1)$.
Iteratively, $\mathbf{z}_{t-1}$ can be obtained by removing noise from $\mathbf{z}_{t}$ conditioned on $\mathbf{y}$.
After the final denoising step, the latent $\mathbf{z}_{0}$ is mapped to generate an image $\widehat{x}$ with the decoder $\mathcal{D}$.

\subsection{\fullmethod (\method)} \label{sec:method}

Given exemplar images that all contain a specific entity, existing subject-driven inversion methods~\cite{Gal:textual-inversion,Kumari:Custom-Diffusion} learn to represent the entity as an identifier token $\mathbf{v} \in \mathbb{R}^d$.
And the learned $\mathbf{v}$ can then be employed in text prompts to produce diverse and novel images, where the entity can be generated with different contexts.
In this paper, we continue the vein of capturing the common action in exemplar images by finding the optimal identifiers.
An overview of our proposed \method is illustrated in \cref{fig:method}.

\unipara{Expanding Semantic Inversion.}
To overcome the preference to low-level appearance features, we apply layer-wise identifier tokens to increase the accommodation of various features.
Specifically, for the $l$-th layer where $l \in [1, L]$ and $L$ is the number of cross-attention layers in the T2I model, a new identifier token $\mathbf{v}_l \in \mathbb{R}^d$ is initialized.
Feeding the prompt with $\mathbf{v}_l$ into the text encoder, the output tokens $\mathbf{y}_l$ control the update of the latents in the $l$-th layer, thus influencing the generation of the visual content.
And the learned tokens from all layers can form a token set $\mathcal{V}$, which can then be paired with different subjects for generation.
Rather than having a single identifier token take on the responsibility of reconstruction, having separate identifiers at different layers effectively ensures that more features are converted, including the action-related features we care about.

\unipara{Learning Gradient Mask with Context-Different Pair.}
The next step is to prevent the identifiers from inverting features that are not relevant to the action and thus contaminating the subsequent image generation.
Given $x^{(a, c)} \in \mathcal{X}$ as an \textit{anchor} sample, where $a$ denotes the specific action, and $c$ denotes the action-agnostic context contained in the image including human appearance and background, we can randomly sample another image $x^{(a, \overline{c})}$ from $\mathcal{X}$, where $\overline{c}$ represents that the context is different from $c$.
Taking the context-different pair $x^{(a, c)}$ and $x^{(a, \overline{c})}$ as the input, we can calculate two gradients of the denoising loss $\mathcal{L}$ with respect to the identifier token $\mathbf{v}$:
{\setlength\abovedisplayskip{2mm}
\setlength\belowdisplayskip{2mm}
\begin{align}
\mathbf{g}^{(a, c)} = \frac{\partial \mathcal{L}^{(a, c)}}{\partial \mathbf{v}}, \\
\mathbf{g}^{(a, \overline{c})} = \frac{\partial \mathcal{L}^{(a, \overline{c})}}{\partial \mathbf{v}}.
\end{align}}%
Note that the subscript $l$ is omitted for the sake of uniformity and clarity.
Each identifier token contains multiple channels, each carrying semantically distinct and independent information.
And the gradient consistency of a channel indicates that the channel is likely to carry information about the specific action.
Therefore, we calculate the absolute value of the difference between the two gradients:
{\setlength\abovedisplayskip{1mm}
\setlength\belowdisplayskip{2mm}
\begin{align}
\bigtriangleup \mathbf{g}^{\overline{c}} = | \mathbf{g}^{(a, c)} - \mathbf{g}^{(a, \overline{c})} |,
\end{align}}%
where the semantic channels with a small difference can be regarded as \textbf{action-related} channels of the action $a$, which are expected to be \textbf{preserved}.
Specifically, we sort the difference from the largest to the smallest, 
taking the value at $\beta$\ percent $\gamma^{\beta}$ as a threshold.
In other words, $\beta$\% of the channels are masked.
Then, the mask that shares the same dimension as $\mathbf{v}$ can be calculated.
For the $k$-th channel,
{\setlength\abovedisplayskip{2mm}
\setlength\belowdisplayskip{2mm}
\begin{align}
\mathbf{m}^{\overline{c}}_k = 
\left\{ 
    \begin{array}{lc}
        0, & \bigtriangleup \mathbf{g}^{\overline{c}}_k \geqslant \gamma^{\beta} \\
        1, & \bigtriangleup \mathbf{g}^{\overline{c}}_k < \gamma^{\beta}\\
    \end{array}.
\right.
\end{align}}%
By overwriting the mask to the gradient of the anchor sample, the action-related knowledge is preserved and incorporated into the update of $\mathbf{v}$, while the updates on action-agnostic channels are ignored.
Note that since the specific visual invariance about the action changes slightly depending on the sample pair, the masked channels may not be exactly the same each time.
Furthermore, both samples use the prompt of the anchor sample $x^{(a, c)}$ when calculating the gradients.
Since the visual context of $x^{(a, \overline{c})}$ is inconsistent with the description in the prompt, the reconstruction loss favours larger gradients in the context-related channels.
In this way, the action-related channels found through the threshold will be more accurate.

\begin{figure*}[!t]    %
  \centering
   \includegraphics[width=\linewidth]{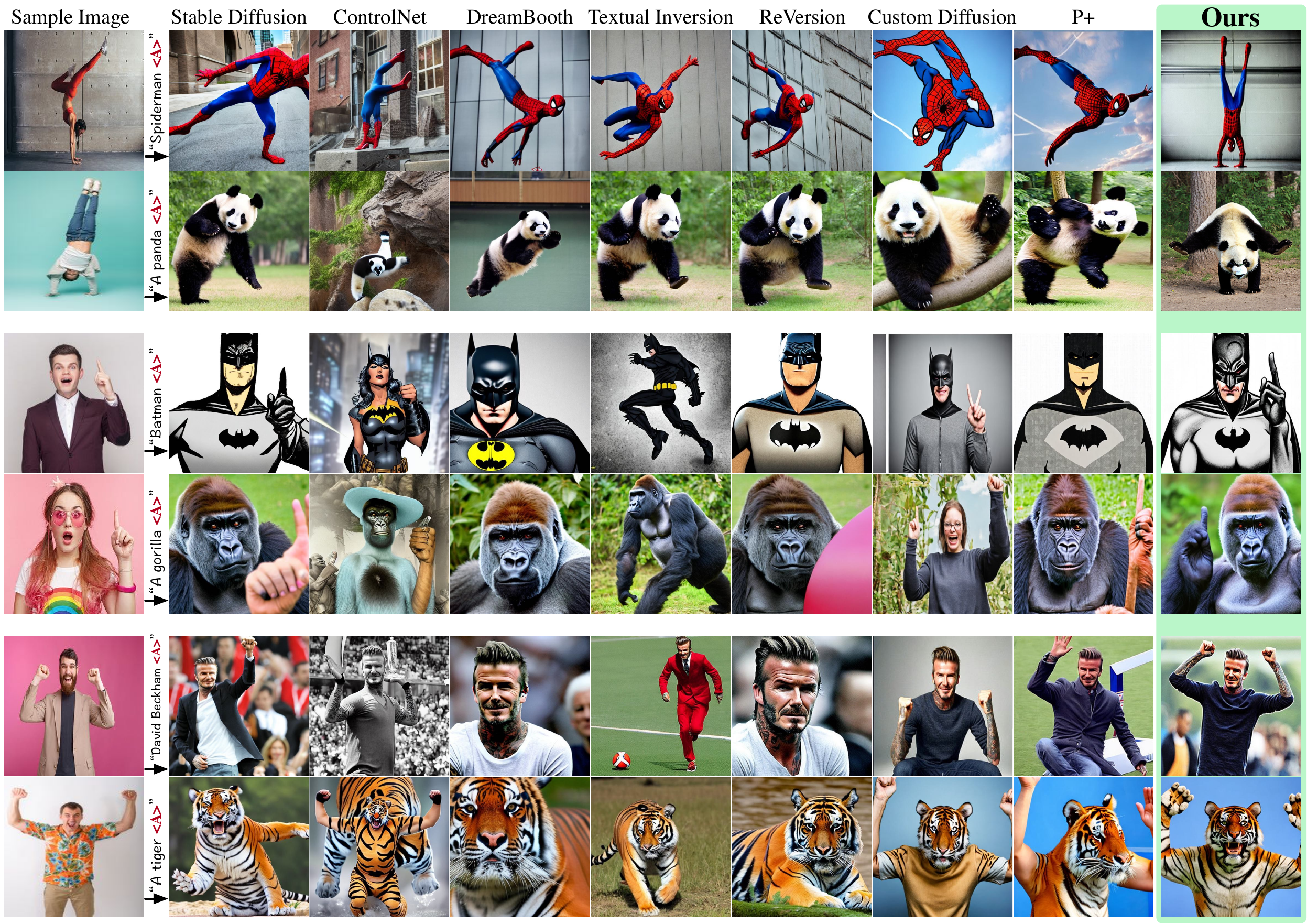}
   \vspace{-6mm}
   \caption{
   \textbf{Visual comparisons of all methods.}
   For each action, we present the generated results showcasing its pairing with a human character and an animal.
   }
         \vspace{-4mm}
   \label{fig:baselines}
\end{figure*}

\unipara{Learning Gradient Mask with Action-Different Pair.}
Although the context-different pairs have the same action semantics, there may be differences in the visualization of the actions, and therefore the channels associated with the most representative action features do not necessarily have a smaller gradient difference.
Since learning the gradient mask with the context-different only pair is not stable and effective enough, we also construct action-different pairs to generate the gradient mask from another perspective.
For each sample $x^{(a, c)}$ in $\mathcal{X}$, we can use it to quickly train a subject-driven customization model (\textit{e.g.}, DreamBooth) that effectively inverts the most of the low-level context information.
Therefore, by filling the prompt template of $x^{(a, c)}$ with action descriptions that are different from $a$, the trained customization model can generate various action images as $\mathcal{X}^{(\overline{a}, c)}$.
Note that this step is not necessary if the users can compile a dataset of varied actions by the same individual using pre-captured images.
However, the data collection is usually arduous and lengthy, making fast training of a subject-driven customization a more convenient solution.
Due to the one-shot training and the concise text,
the generated images $\mathcal{X}^{(\overline{a}, c)}$ may not be consistent with the action descriptions, or the context may differ from the original $x^{(a, c)}$, but in practice, we have found that the quality is sufficient to diversify the action variation.
In this way, when $x^{(a, c)}$ is sampled during training, we can randomly sample a image $x^{(\overline{a}, c)}$ from $\mathcal{X}^{(\overline{a}, c)}$ to construct the action-different pair.
And the gradient of $x^{(\overline{a}, c)}$ with respect to the token $\mathbf{v}$ can be calculated as
{\setlength\abovedisplayskip{2mm}
\setlength\belowdisplayskip{2mm}
\begin{align}
\mathbf{g}^{(\overline{a}, c)} = \frac{\partial \mathcal{L}^{(\overline{a}, c)}}{\partial \mathbf{v}}.
\end{align}}%
Similarly, both samples use the prompt of the anchor sample $x^{(a, c)}$.
We can also calculate the absolute value of the gradient difference:
{\setlength\abovedisplayskip{2mm}
\setlength\belowdisplayskip{2mm}
\begin{align}
\bigtriangleup \mathbf{g}^{\overline{a}} = | \mathbf{g}^{(a, c)} - \mathbf{g}^{(\overline{a}, c)} |,
\end{align}}%
where the semantic channels with small difference can be regarded as \textbf{context-related} channels of the action $a$, which are expected to be \textbf{masked}.
Therefore, we have
{\setlength\abovedisplayskip{2mm}
\setlength\belowdisplayskip{2mm}
\begin{align}
\mathbf{m}^{\overline{a}}_k = 
\left\{ 
    \begin{array}{lc}
        0, & \bigtriangleup \mathbf{g}^{\overline{a}}_k < \lambda^{\beta} \\
        1, & \bigtriangleup \mathbf{g}^{\overline{a}}_k \geqslant \lambda^{\beta}\\
    \end{array},
\right.
\end{align}}%
where $\lambda^{\beta}$ is the threshold here to mask $\beta$\% of the channels.

\unipara{Merging Gradient Masks for Context.}
Due to the noise introduced by context variations, identifying action-relevant channels using only context-different or action-different pairs would be difficult and unreliable.
As an evidence, the average overlap rate of channels preserved by both masks at each training step is 30.26\%.
Therefore, given the input triple $\mathcal{I} = \left\{ x^{(a, \overline{c})}, x^{(a, c)}, x^{(\overline{a}, c)} \right\}$, we can merge the two obtained masks $\mathbf{m}^{\overline{a}}$ and $\mathbf{m}^{\overline{c}}$ to get the final context mask $\mathbf{m}$.
In practice, we keep only the intersection of the unmasked channels as unmasked, as we find this merging strategy performs better.
Formally, we have
{\setlength\abovedisplayskip{2mm}
\setlength\belowdisplayskip{1mm}
\begin{align} \label{eq:mask_merging}
\mathbf{m} = \mathbf{m}^{\overline{c}} \cap {\mathbf{m}}^{\overline{a}},
\end{align}}%

\noindent which is overwritten to the gradient of the anchor sample:
{\setlength\abovedisplayskip{-2mm}
\setlength\belowdisplayskip{0mm}
\begin{align}
\widetilde{\mathbf{g}} ^{(a, c)} = \mathbf{m} \odot \mathbf{g}^{(a, c)}.
\end{align}}%

\noindent Note that the masked gradient $\widetilde{\mathbf{g}} ^{(a, c)}$, where action-agnostic channels are considered to be masked, is the only gradient used to update $\mathbf{v}$.
Therefore, our identifiers can adequately invert action-related features.
\section{Experiments}

\subsection{Experiment Setup}

\unipara{Baselines.}
For the baselines included in the comparison, we select Stable Diffusion~\cite{Rombach:Stable-Diffusion}, ControlNet~\cite{Zhang:ControlNet}, DreamBooth~\cite{Ruiz:DreamBooth}, Textual Inversion~\cite{Gal:textual-inversion}, ReVersion~\cite{Huang:ReVersion}, Custom Diffusion~\cite{Kumari:Custom-Diffusion} and P+~\cite{VoynovL:P-Plus}.

\unipara{Implementation Details.}
For \method, we set the masking ratio $\beta$ to 0.6 and use the AdamW~\cite{Loshchilov:AdamW} optimizer with a learning rate of 2e-4, while the training takes 3000 steps.
For efficiency, DreamBooth for action-different pairs does not generate class-preservation images. While only one image is used for training, the initial learning rate is 1e-6, and the training takes 2000 steps.
For a fair comparison, we use 50 steps of the DDIM~\cite{Song:DDIM} sampler with a scale of 7.5 for all methods.
Unless otherwise specified, Stable Diffusion v2-1-base is selected as the default pre-trained model, and images are generated at a resolution of 512$\times$512.
All experiments are conducted on a NVIDIA A100 GPU.

\begin{figure*}[!t]    %
  \centering
   \includegraphics[width=0.78\linewidth]{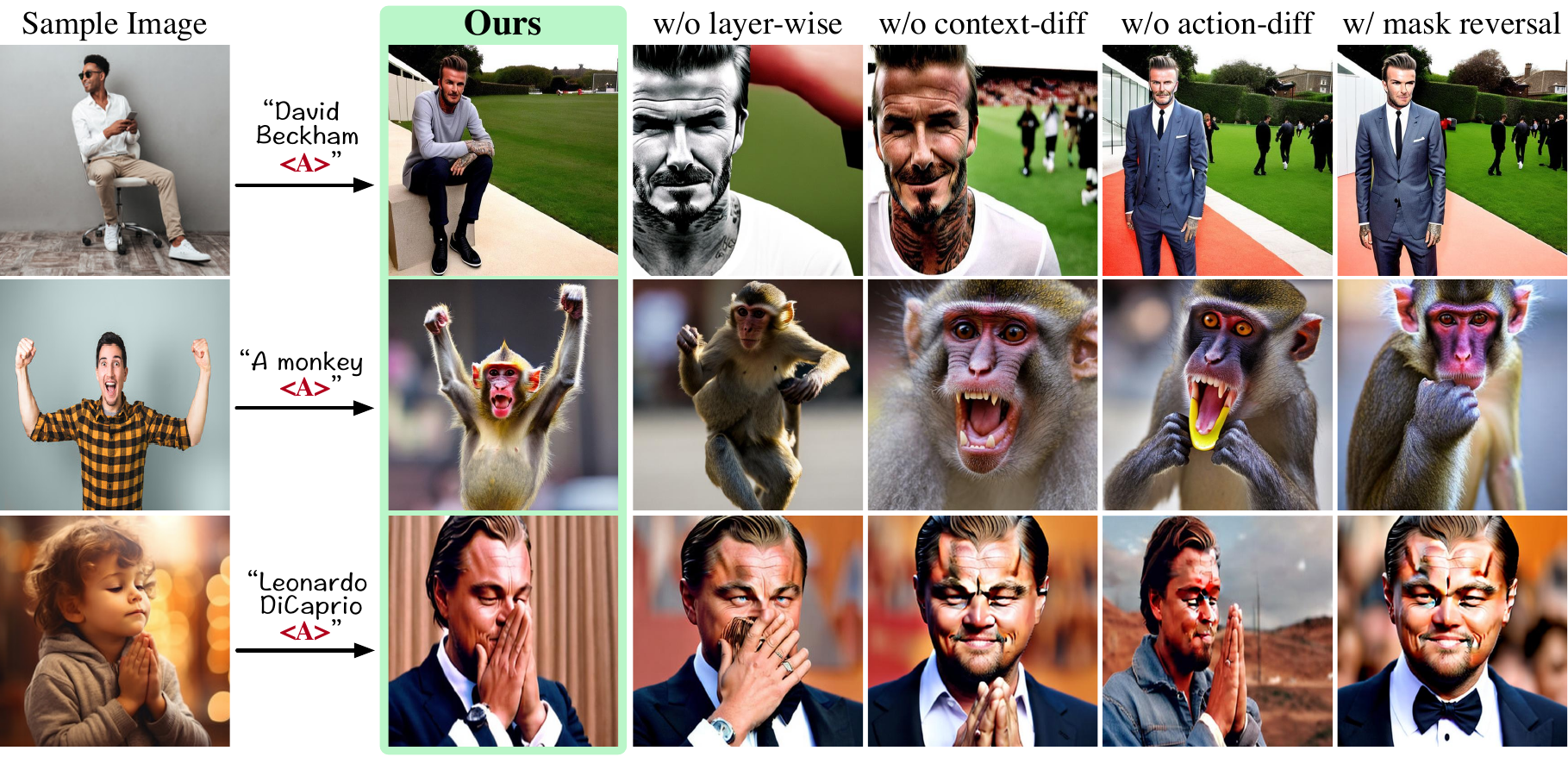}
   \vspace{-2mm}
   \caption{
   \textbf{Ablation study.}
   We remove or revise one implementation at a time to demonstrate the effects of the identifier extension and the gradient masking.
   }
         \vspace{-4mm}
   \label{fig:ablation}
\end{figure*}

\subsection{Quantitative Comparison}
We perform the quantitative comparison with human evaluators to assess the quantitative performance.
For each subject-action pair, four images are randomly sampled from the images generated by different methods.
Given (1) the exemplar images of a specific action, and (2) the textual name of the subjects, human evaluators are asked to determine whether (1) the generated action is consistent with those in the exemplar images, and (2) the generated character corresponds with the name without obvious deformations, defects, or abnormalities.
A generated image will only be considered totally correct if both the action and the character are correctly generated.

\begin{table}[!t]
\tablestyle{5pt}{1.0}
\setlength\tabcolsep{4pt}
\def\w{20pt} 
    \caption{%
  \textbf{Quantitative comparisons with competing methods.} 
  Action, subject and total accuracies (\%) are reported.
}
\vspace{-2mm}
\scalebox{1.1}{
    \begin{tabular}{l|c|c|c}
    \shline
    \textbf{Methods} & \textbf{Action} & \textbf{Subject} & \textbf{Total} \\
    \shline
    Stable Diffusion~\cite{Rombach:Stable-Diffusion} & 30.71 & 84.51 & 27.17 \\
    ControlNet~\cite{Zhang:ControlNet} & 41.30 & 42.66 & 19.29 \\
    DreamBooth~\cite{Ruiz:DreamBooth} & 2.45 & \textbf{95.65} & 2.45 \\
    Textual Inversion~\cite{Gal:textual-inversion} & 2.17 & 86.14 & 1.90 \\
    ReVersion~\cite{Huang:ReVersion} & 1.63 & 84.51 & 1.63 \\
    Custom Diffusion~\cite{Kumari:Custom-Diffusion} & 29.62 & 53.53 & 7.07 \\
    P+~\cite{VoynovL:P-Plus} & 26.90 & 80.16 & 20.92 \\
    \rowcolor[rgb]{ .949,  .949,  .949} \textbf{\method (Ours)}  & \textbf{60.33} & 85.87 & \textbf{51.09} \\
    \shline
    \end{tabular}%
    }
\vspace{-5mm}
  \label{tab:main_results}%
\end{table}%

\cref{tab:main_results} reports the action, subject and total accuracy for all methods.
Some observations are worth highlighting:
(1) Given the textual descriptions of the actions, Stable Diffusion yields the highest total accuracy of all baseline methods.
This suggests that the existing baselines do not take full advantage of the exemplar images.
(2) Despite relying on the skeleton as the condition to improve the action generation, ControlNet fail to maintain the performance of subject generation, resulting in an unsatisfactory total accuracy.
(3) The action accuracy of DreamBooth, Textual Inversion, and ReVersion is incredibly low, reflecting their complete failure to invert the action-related features.
(4) Custom Diffusion and P+ improve action accuracy at more or less the expense of subject accuracy.
(5) Attribute to the extended semantic conditioning space and the gradient masking strategy, our \method dramatically improves the accuracy of action generation while maintaining excellent subject accuracy.
As a result, \method achieves the best total accuracy, outperforming the baselines by 23.92\%.

\begin{figure*}[!t]    %
  \centering
   \includegraphics[width=0.78\linewidth]{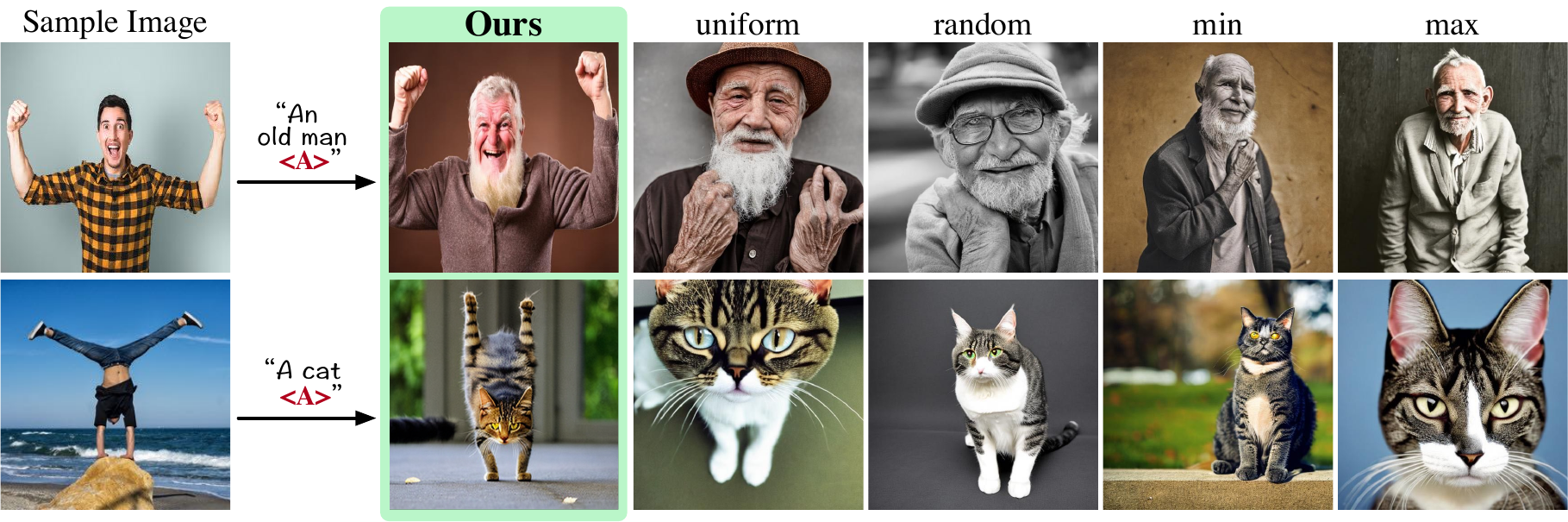}
   \vspace{-2mm}
   \caption{
   \textbf{Visual comparison of masking strategies.}
   The four compared strategies fail to mask the updates from agnostic channels and invert the action-related features.
   }
         \vspace{-4mm}
   \label{fig:mask_analysis}
\end{figure*}

\subsection{Qualitative Comparison}
\cref{fig:baselines} illustrates the qualitative comparison of all methods involved.
It can be observed that although text descriptions of the actions are provided, the actions generated by Stable Diffusion still differ from the examples.
ControlNet can only maintain a rough consistency in posture and struggles to match the generated subjects to the desired requirements, resulting in incomplete or distorted body structures, while sacrificing diversity.
And the subject-driven customization methods, as discussed earlier, fail to generate the actions or exhibit appearance characteristics that differ from the specified subjects.
This suggests that they are unable to convert only the features associated with the actions.
Giving the credit to the design from a perspective of gradient, our \method decouples action-related features from action-agnostic information and blocks the inversion of the latter.
This allows \method to effectively model the invariance of the action and transfer it to different characters and animals without sacrificing image quality and variety.

\subsection{Ablation Study}
We conduct ablation experiments on \benchmark to verify the individual effects of the proposed contributions.
From the generation results in \cref{fig:ablation}, it can be observed that
(1) The removal of the extension to the semantic conditioning space diminishes the inversion ability of \method. 
(2) Both the gradient masks learned from the context-different and the action-different pairs are essential.
Removing either one can lead to inadequate learning of action knowledge or a degradation in the quality of the subject's appearance.
We attribute this to the fact that learning from a single pair is inherently noisy due to varied action visuals and the interference of action-irrelevant information.
(3) We also attempt to reverse the gradient masks, \textit{i.e.}, updates to channels that should have been masked are preserved, and updates to other channels are cancelled.
Obviously, this will result in action-related features not being inverted.

\subsection{Further Analysis}

\unipara{Impact of Masking Strategy.} 
To validate the masking strategy in our \method, we compare it with four other strategies in \cref{fig:mask_analysis}.
Specifically, on the gradients for each update:
(1) Uniform: we uniformly mask $\beta$ percent of channels.
(2) Random: we randomly mask $\beta$ percent of channels.
(3) Min: we mask $\beta$ percent of channels with the lowest value.
(4) Max: we mask $\beta$ percent of channels with the highest value.
We observe that none of these four strategies successfully captures high-level features related to actions, since the images they generate are independent of the specified action.
And the comparison also shows that the effectiveness of our \method not only depends on the masking itself, but also requires learning action-agnostic channels by modeling the invariance of action and context.

\begin{figure}[!t]    %
  \centering
   \includegraphics[width=\linewidth]{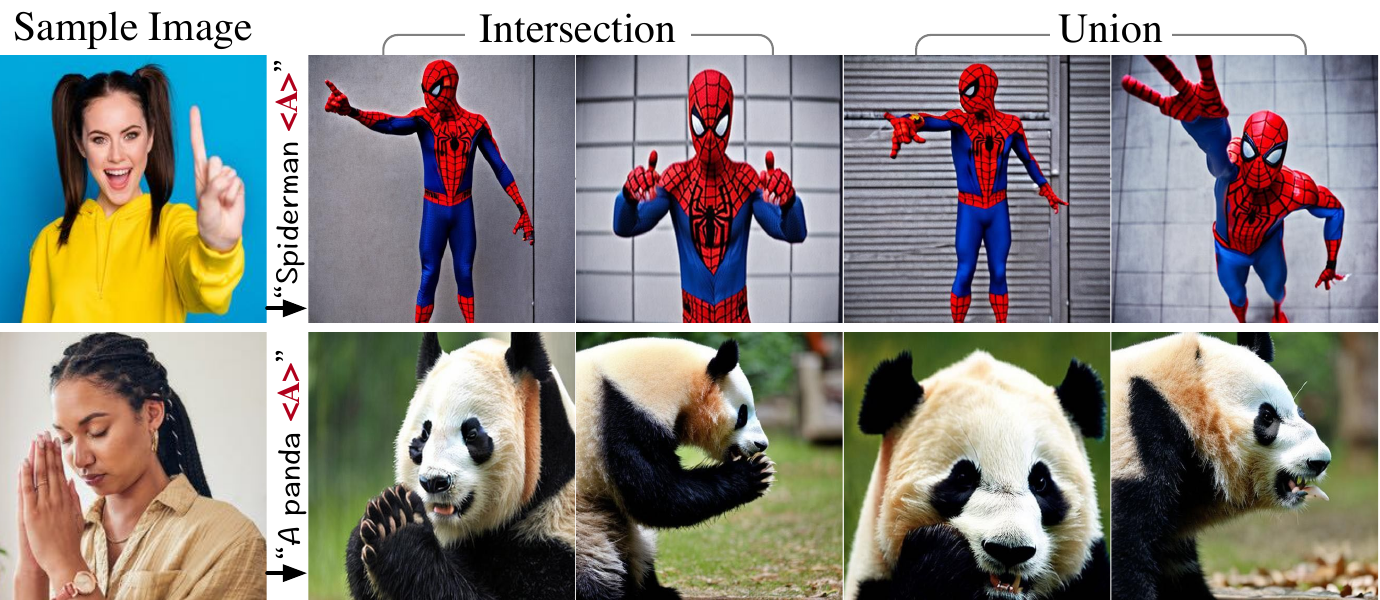}
   \vspace{-5mm}
   \caption{
   \textbf{Effect of gradient mask merging strategy.}
   Preserving the intersection of gradient masks can better invert the representative action features.
   }
   \vspace{-4mm}
   \label{fig:hyper_mask_merging}
\end{figure}

\unipara{Impact of Gradient Mask Merging Strategy.} 
As shown in \cref{eq:mask_merging}, \method takes the intersection of the two gradient masks as the default merging strategy.
We compare this with selecting the union of the two masks, and illustrate the generation results in \cref{fig:hyper_mask_merging}.
Since only channels that are preserved on both masks are updated, taking the intersection can effectively filter out action-agnostic features, leading to better customization of the actions.
In contrast, taking the union may dilute the most representative action features due to the preserved context information.

\begin{figure}[!t]    %
  \centering
   \includegraphics[width=\linewidth]{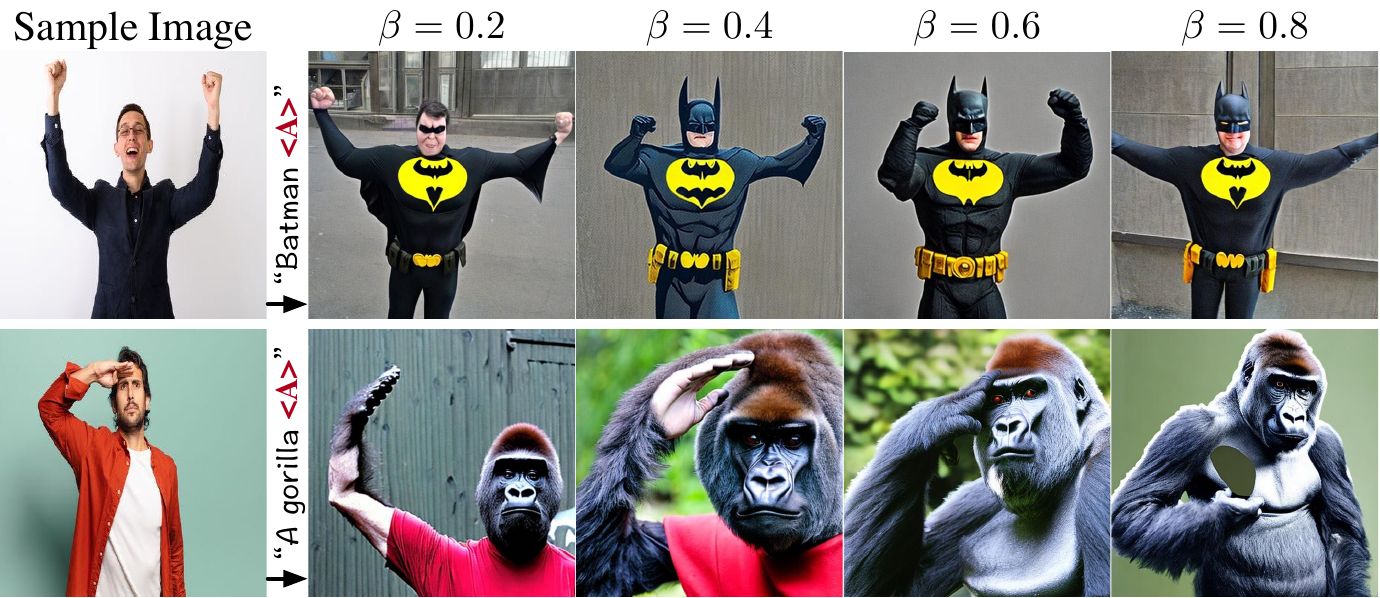}
   \vspace{-5mm}
   \caption{
   \textbf{Effect of the masking ratio $\beta$.}
   A value close to 0.5 can balance the inversion of action-related features and the removal of action-agnostic features.
   }
   \vspace{-4mm}
   \label{fig:hyper_beta}
\end{figure}

\unipara{Impact of Masking Ratio $\beta$.} 
In \cref{fig:hyper_beta}, we vary the masking ratio $\beta$ from 0.2 to 0.8.
When $\beta$ is small, fewer dimensions of the gradient are masked, and more action-agnostic features are retained to hinder the generation of the subject's appearance.
This situation improves as $\beta$ is gradually increased.
However, when $\beta$ is relatively large, due to the large number of masked dimensions, some of the most discriminative features of actions may not be inverted, resulting in incomplete learning of actions.
Note that the optimal value of $\beta$ may be different for different actions.

\section{Conclusion}

In this paper, we investigate an under-explored text-to-image generation task, namely action customization.
To understand the challenge of the task, we first visualize the inadequacy of existing subject-driven methods in extracting action-related features from the entanglement of action-agnostic context features.
Then, we propose a novel method named \method to learn action-specific identifiers from the given images.
To increase the accommodation of knowledge relevant to the action,
\method extends the inversion process with layer-wise identifier tokens.
Furthermore, \method generates gradient masks to block the contamination of action-agnostic features at the gradient level.
We also contribute the \benchmark for evaluating performance on the task.
Since there is a growing need to synthesize action-specific images with various new subjects,
we hope that our work can highlight this important direction.

\noindent \textbf{Acknowledgement}
This work was supported by STI 2030—Major Projects (2022ZD0208800), NSFC General Program (Grant No. 62176215).
This work was supported by Alibaba Group through Alibaba Research Intern Program.

\clearpage
{
    \small
    \bibliographystyle{ieeenat_fullname}
    \bibliography{ref}
}

\clearpage
\setcounter{page}{1}
\maketitlesupplementary

\appendix
\renewcommand{\thesection}{\Alph{section}}

\section{Benchmark Details}

In this section, we describe the presented \textbf{\benchmark} in detail.
The full benchmark will be publicly available.

\subsection{Actions}

We define eight diverse, unique and representative actions as follows:

\begin{itemize}
\item \textbf{salute}: ``\textit{salutes}''
\item \textbf{gesture}: ``\textit{raises one finger}''
\item \textbf{cheer}: ``\textit{raises both arms for cheering}''
\item \textbf{pray}: ``\textit{has hands together in prayer}''
\item \textbf{sit}: ``\textit{sits}''
\item \textbf{squat}: ``\textit{squats}'' 
\item \textbf{meditate}: ``\textit{meditates}''
\item \textbf{handstand}: ``\textit{performs a handstand}''
\end{itemize} %

\noindent where the action categories (displayed in \textbf{boldface}) are used only to distinguish between actions, and the actions can be best described with the exemplar images.
And the text descriptions (displayed in \textit{italics}) that are used for Stable Diffusion are obtained using an image captioning model.

\subsection{Subjects}

We provide 23 subjects for evaluation as follows:

\begin{itemize}
\item \textbf{generic human}: ``\textit{A boy}'', ``\textit{A girl}'', ``\textit{A man}'', ``\textit{A woman}'', ``\textit{An old man}''
\item \textbf{well-known personalities}: ``\textit{Barack Obama}'', ``\textit{Michael Jackson}'', ``\textit{David Beckham}'', ``\textit{Leonardo DiCaprio}'', ``\textit{Messi}'', ``\textit{Spiderman}'', ``\textit{Batman}''
\item \textbf{animals}: ``\textit{A dog}'', ``\textit{A cat}'', ``\textit{A lion}'', ``\textit{A tiger}'', ``\textit{A bear}'', ``\textit{A polar bear}'', ``\textit{A fox}'', ``\textit{A cheetah}'', ``\textit{A monkey}'', ``\textit{A gorilla}'', ``\textit{A panda}''
\end{itemize} %

\noindent where diverse and unseen subjects and the introduction of animals demand that, models not only retain pre-trained knowledge without forgetting, but also accurately generate animal representations without distortion or anomalies.

\section{Baseline Details}    %

All baselines use the prompt template provided by the \benchmark.
Each prompt details its image content, leaving the action blank for filling with identifiers from different methods.
Other details are:

\begin{itemize}
\item \textbf{ControlNet}~\cite{Zhang:ControlNet}: We use OpenPose~\cite{Cao:OpenPose} as a preprocessor to estimate the human pose of the given reference image.
\item \textbf{DreamBooth}~\cite{Ruiz:DreamBooth}: 
The training is with a batch size of 2 and a learning rate of 5e-5. 
The number of training steps is set to 1000, and 50 images are generated for prior preservation.
\item \textbf{Textual Inversion}~\cite{Gal:textual-inversion}: The training is with a batch size of 2 and a learning rate of 2.5e-4. The number of training steps is set to 3000.
\item \textbf{ReVersion}~\cite{Huang:ReVersion}: The training is with a batch size of 2 and a learning rate of 2.5e-4.
The number of training steps is set to 3000.
The weighting factors of the denoising loss and the steering loss are set to 1.0 and 0.01.
The temperature parameter in the steering loss is set to 0.07.
And in each iteration, 8 positive samples are randomly selected from the basis preposition set.
\item \textbf{Custom Diffusion}~\cite{Kumari:Custom-Diffusion}: The training is with a batch size of 2 and a learning rate 1e-5.
The number of training steps is 2000.
And the number of regularization images is 200.
\item \textbf{P+}~\cite{VoynovL:P-Plus}: The training is with a batch size of 8 and a learning rate 5e-3.
The number of training steps is 500.
\end{itemize}

\section{Additional Experimental Results}

\begin{figure}[!t]    %
  \centering
   \includegraphics[width=\linewidth]{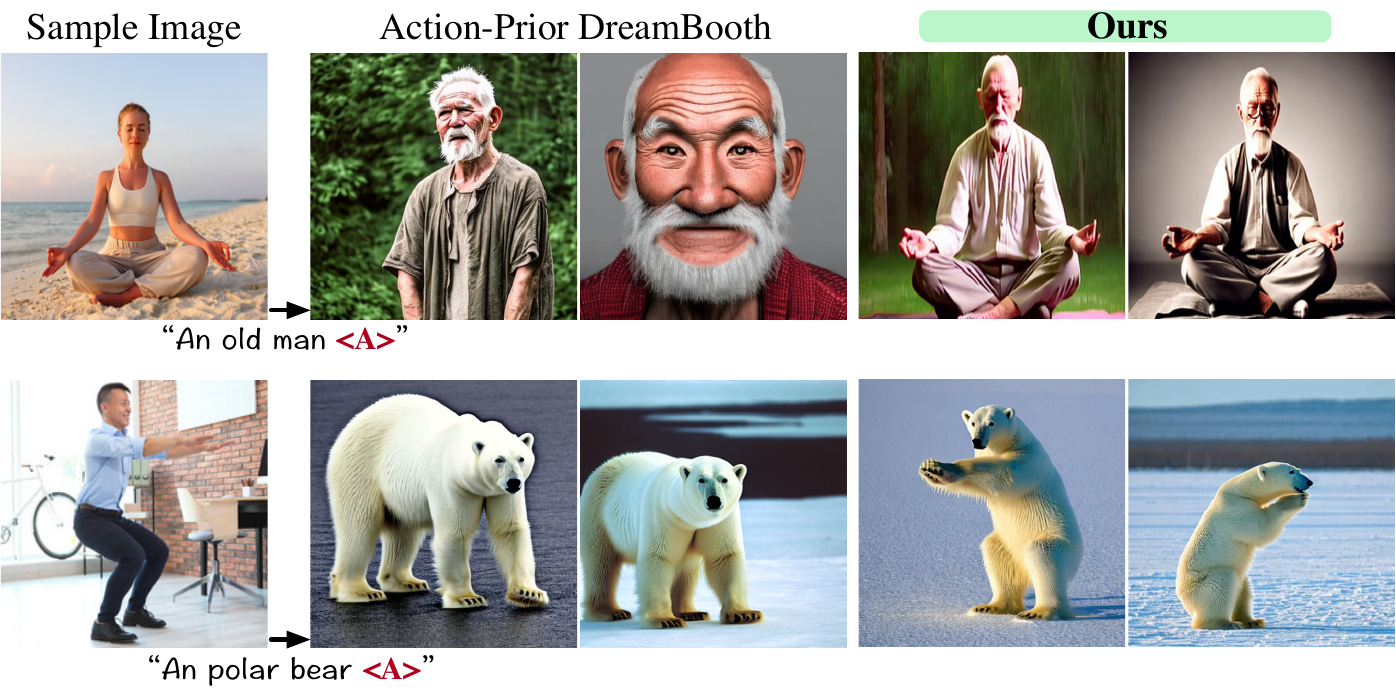}
   \vspace{-4mm}
   \caption{
   \textbf{Comparison with action-prior DreamBooth.}
   This extended DreamBooth still struggles with inverting action features.
   }
   \vspace{-2mm}
   \label{fig:action_prior_dreambooth}
\end{figure}

\subsection{Comparison with Action-Prior DreamBooth}

Our \method utilizes the generated action-different samples with the same context to capture the context-related features.
To analyze the advantages of controlling updates with these data rather than directly employing them in training,
we present a new baseline named action-prior DreamBooth, which replaces the class prior generated by original Stable Diffusion with these action-different samples.
Therefore, in addition to the inherent action invariance, contextual invariance also emerges in the training data.
However, as shown in \cref{fig:action_prior_dreambooth}, this new baseline still struggles with inverting action-specific features.
This observation suggests a lack of ability to capture high-level invariance.

\subsection{Generalization Across Diverse Styles}

\begin{figure}[htbp]    %
  \centering
   \includegraphics[width=\linewidth]{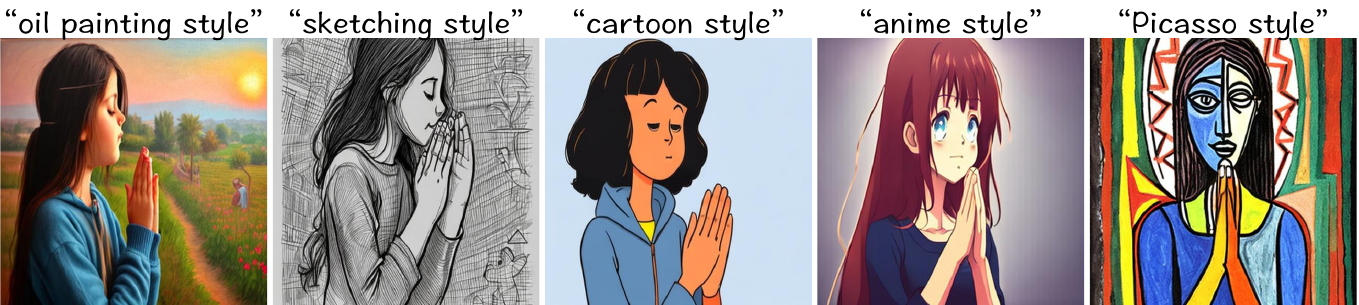}
   \vspace{-4mm}
   \caption{
   \textbf{\method can generate images with different styles by prompting.}
   The original prompt is ``\textit{A girl} $<\!\!\text{A}\!\!>$'' where ``$<\!\!\text{A}\!\!>$'' represents the action \textbf{pray}.
   }
   \vspace{-2mm}
   \label{fig:ADI_with_style}
\end{figure}

\method is designed to separate and inverse abstract the action concepts from the details of subjects and objects, background, color, or style in user images.
This allows the generation images to generalize to specific styles through prompting, shown as \cref{fig:ADI_with_style}.

\subsection{Visualization of Cross-Attention Maps}

\begin{figure}[htbp]    %
  \centering
   \includegraphics[width=\linewidth]{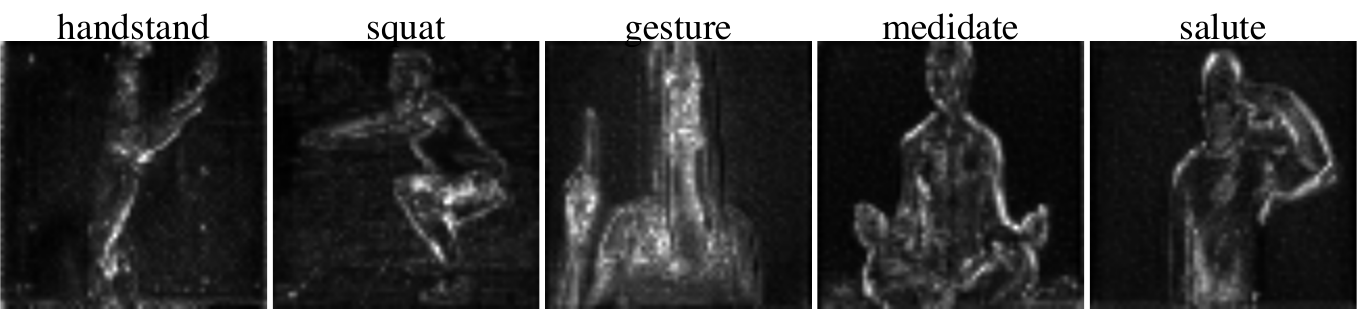}
   \vspace{-4mm}
   \caption{
   \textbf{Visualization of cross-attention maps associated with the learned identifiers.}
   }
   \vspace{-2mm}
   \label{fig:attention_map_v2}
\end{figure}

To explain why certain channels can be interpreted as ``action-related'', we visualize the cross-attention maps related to the learned identifiers in \cref{fig:attention_map_v2}.
As observed, the learned identifiers focus more on the contour information of the actions rather than the human body.
This indicates that \method avoids reversion on appearance information, thereby enabling generalization to different subjects.

\subsection{Visualization of Action-Different Pairs}

\begin{figure}[!t]    %
\vspace{-183mm}
  \centering
   \includegraphics[width=\linewidth]{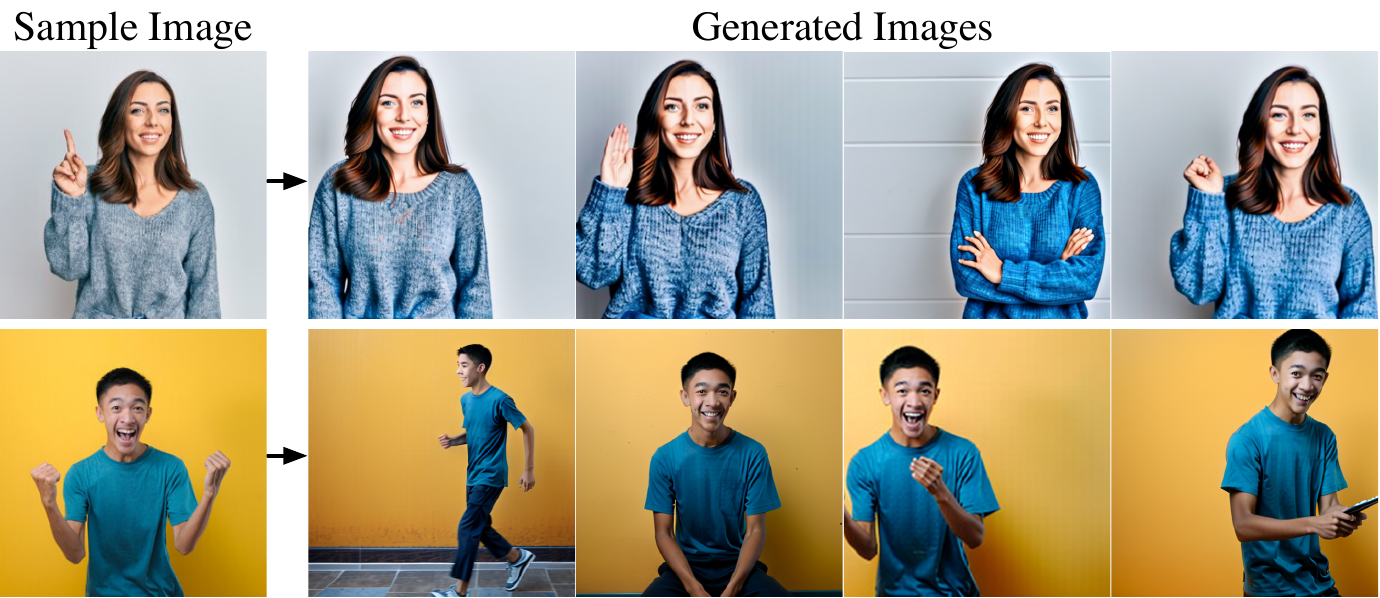}
   \vspace{-4mm}
   \caption{
   \textbf{Visualization of subject-driven generation results for action-different pairs.}
   }
   \vspace{-2mm}
   \label{fig:dreambooth_for_person}
\end{figure}

We present the generated images in the action-different pairs in \cref{fig:dreambooth_for_person} for reference.
Using only a single image for training, the subject-driven model can change actions while preserving contextual information as much as possible. Although the quality of the image may be insufficient, it does not hinder the final inversion of action knowledge.

\subsection{Additional Qualitative Results}

\begin{figure*}[!t]    %
  \centering
   \includegraphics[width=0.85\linewidth]{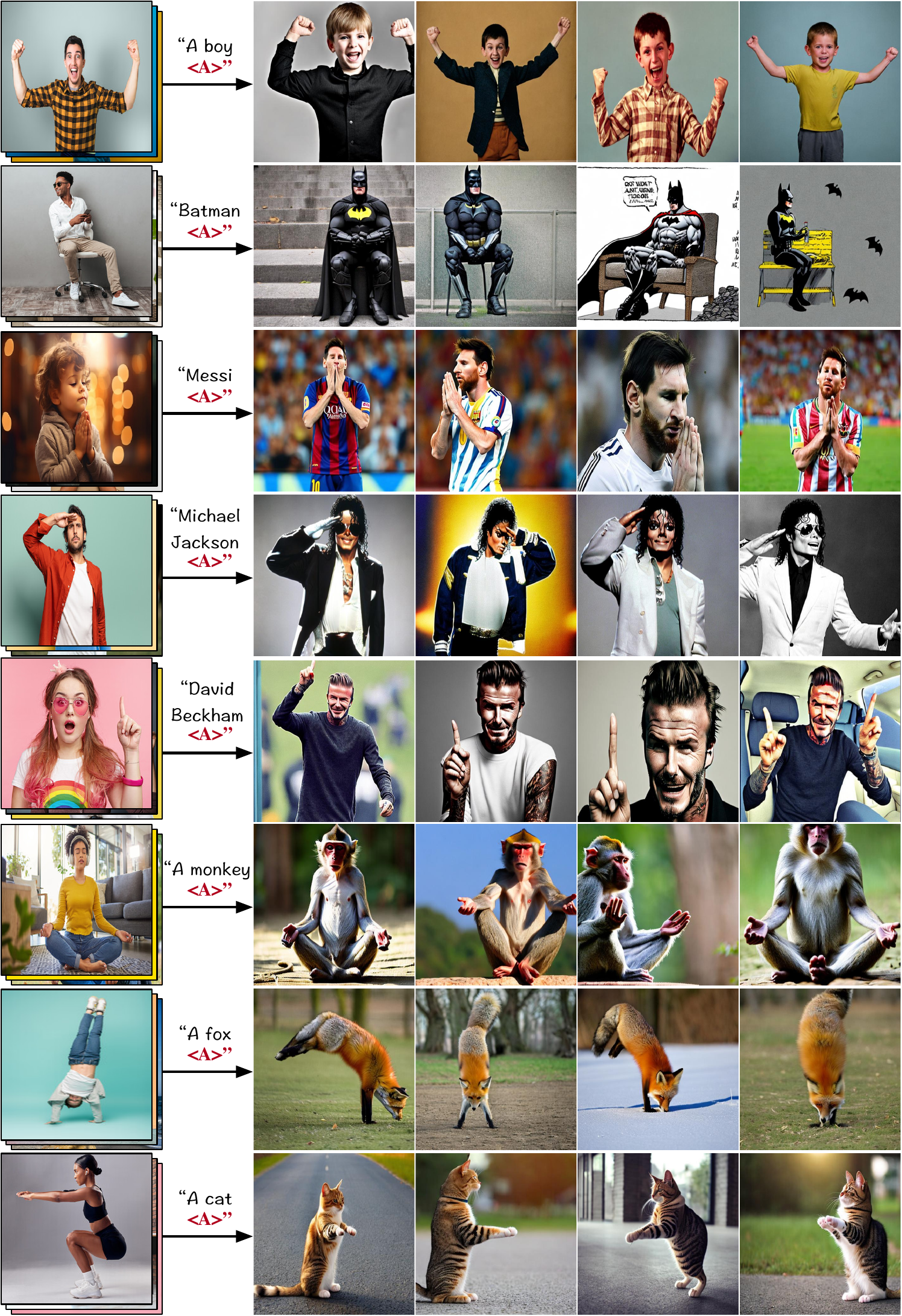}
   \caption{
   \textbf{Additional generation results by \method, encompassing all the actions within \benchmark.}  %
   }
         \vspace{-4mm}
   \label{fig:additional_results}
\end{figure*}

To show the effectiveness of \method,
we illustrate additional generation results in \cref{fig:additional_results}, covering all actions within \benchmark.
The generated images maintain the same action while offering a rich diversity, indicating that the learned identifiers contain solely action information and do not encapsulate irrelevant contextual details such as background, appearance, or even orientation.

\break

\end{document}